# Investigation of Customized Medical Decision Algorithms Utilizing Graph Neural Networks

Yafeng Yan[1], Shuyao He[2], Zhou Yu[3], Jiajie Yuan[4], Ziang Liu[5], Yan Chen[6]

[1]Stevens Institute of Technology,USA,yanyafeng0105@gmail.com

[2]Northeastern University,USA,he.shuyao@northeastern.edu

[3]University of Illinois at Chicago,USA,zyu941112@gmail.com

[4]Brandeis University,USA,jiajieyuan@brandeis.edu

[5]Carnegie Mellon University,USA,ziangliu@alumni.cmu.edu

[6]Stevens Institute of Technology,USA,yanchen@alumni.stevens.edu

*Abstract*—Aiming at the limitations of traditional medical decision system in processing large-scale heterogeneous medical data and realizing highly personalized recommendation, this paper introduces a personalized medical decision algorithm utilizing graph neural network (GNN). This research innovatively integrates graph neural network technology into the medical and health field, aiming to build a high-precision representation model of patient health status by mining the complex association between patients' clinical characteristics, genetic information, living habits. In this study, medical data is preprocessed to transform it into a graph structure, where nodes represent different data entities (such as patients, diseases, genes, etc.) and edges represent interactions or relationships between entities. The core of the algorithm is to design a novel multi-scale fusion mechanism, combining the historical medical records, physiological indicators and genetic characteristics of patients, to dynamically adjust the attention allocation strategy of the graph neural network, so as to achieve highly customized analysis of individual cases. In the experimental part, this study selected several publicly available medical data sets for validation, and the results showed that compared with traditional machine learning methods and a single graph neural network model, the proposed personalized medical decision algorithm showed significantly superior performance in terms of disease prediction accuracy, treatment effect evaluation and patient risk stratification.

*Keywords—graph neural network, multi-scale fusion mechanism, medical atlas, attention mechanism*

## I. INTRODUCTION

In today's medical and health field, how to effectively use massive and complex medical data to provide more accurate and personalized medical decision support for each patient has become a key issue to be solved. Traditional medical decision-making system often relies on the doctor's experience judgment and limited statistical analysis, it is difficult to fully capture the huge differences among patients and the complex correlation among medical data. Therefore, exploring a new method that can deeply tap the potential value of medical data and realize efficient personalized medical decision-making is essential.

Graph Neural Networks [1] (GNNs), as a powerful machine learning tool, have received widespread attention for their excellent performance in processing non-Euclidean data, especially complex network structure data. Through information dissemination and aggregation operations on the graph structure, GNN can learn the deep feature representation of nodes and their neighbors, which provides a powerful means to understand the complex interactions between various entities in the medical field (such as patients, diseases, drugs, genes, etc.)[2-5].

This paper aims to explore and realize intelligent and personalized medical decisions through the following core contents: First, we detail how to transform diverse medical data (including but not limited to electronic medical records, genomic data, medical image reports, lifestyle information, etc.) into a unified graphical representation. This transformation requires not only the preservation of the rich information of the original data, but also the effective encoding of multiple associations between entities, such as the association of diseases and symptoms, and the association of gene mutations and disease risk. Secondly, considering the particularity of medical data, this study will explore and select GNN models suitable for the characteristics of the medical field, such as convolutional network (GCN) and Attention network [6-8] (GAT), etc., and optimize and adjust them according to the specific needs of medical decision-making. Emphasis is placed on designing model structures when dealing with highly heterogeneous medical data. By designing a novel feature learning mechanism, this study aims to extract high-dimensional feature representations from the graph structure that reflect patients' personalized medical needs. Combined with the medical knowledge graph and deep learning technology, algorithms will provide in-depth analysis of each patient's condition, providing customized disease prediction, treatment recommendations, and risk assessment.

We will conduct extensive experiments based on real-world medical data sets and compare them with existing methods to evaluate performance improvements in disease diagnosis accuracy, treatment effect prediction, and patient risk stratification. At the same time, the application potential of the algorithm in practical medical scenarios is discussed, including auxiliary clinical decision making and personalized health management plan making. To sum up, this study aims to build an efficient and accurate personalized medical decision support system through the powerful capability of graph neural network, promote the development of medical care in a more intelligent and personalized direction, and provide solid technical support for realizing the vision of "precision medicine".

## II. RELATED WORK

In the scientific realm of personalized medical decision-making, remarkable progress has been made in recent years, laying a solid foundation for improving the precision and personalized level of health care. This section will comprehensively examine the evolution of medical decision algorithm, the innovation breakthrough of graph neural network technology, and the cutting-edge research results achieved by the integration of the two in the field of personalized medical decision making.

Evolution and breakthrough of medical decision algorithm: Medical decision algorithm, as the product of the cross integration of mathematics, statistics and computer science, its core lies in the use of advanced algorithms to extract key information from the complicated medical data to guide medical practice[9-11]. In the early days, traditional methods based on statistics, such as logistic regression and support vector machines[12], played a fundamental role in disease prediction and medical resource management. However, the limited ability of these methods to deal with highly complex, non-linear medical data relationships and to mine individualized features has led researchers to seek more advanced analytical tools.

As a revolutionary technology, graph neural networks (GNNs) are unique in that they can learn directly on the graph data structure, and achieve a deep understanding of the internal structure of the data by capturing complex interactions and information flows between nodes. GNN has not only made remarkable achievements in the fields of computer vision[13-15],3d-aware[16-17], linguistic system[18-19], etc., but also shown great potential in the field of medical health, especially in medical image recognition, disease association network analysis, etc. Its powerful graph structure representation learning ability brings new perspectives and tools to medical decision-making.

With the popularity of the concept of precision medicine, the research of personalized medical decision algorithms has become a focal point. These algorithms are dedicated to integrating patients' clinical parameters, genetic background, lifestyle, and other dimensions of information, to tailor the diagnosis and treatment plan for each patient through sophisticated data analysis. Currently, innovative methods such as medical image analysis based on deep learning and personalized treatment path planning based on reinforcement learning are propelling medical decision science to new heights. Notably, deep learning techniques have significantly enhanced the capabilities of medical image segmentation, as demonstrated by Zi et al. (2024), who explored the application of these methods in both segmentation and 3D reconstruction of medical images[20]. By deeply integrating graph neural networks with personalized medical decision algorithms, scholars aim to overcome the limitations of current models and achieve fine modeling and prediction of individual health conditions. This integration promises to unlock new potentials in precision medicine, fostering more accurate and personalized healthcare solutions.

Among them, personalized medical decision algorithms are an important research area, aiming to provide customized diagnosis and treatment plans for each patient based on their individual characteristics and medical data. In this area, many approaches based on deep learning have been proposed in recent years, alongside new technologies that utilize graph neural networks to process medical data. Traditional medical decision algorithms usually adopt statistical methods or machine learning methods. These methods, while performing well in some situations, struggle to capture complex relationships and individual characteristics in medical data. Therefore, more and more research has begun to explore new methods, such as deep learning-based medical image analysis and personalized treatment decision-making. Notably, Yan et al. (2024) have applied neural networks to enhance the accuracy of survival predictions across diverse cancer types, illustrating the potential of these advanced computational techniques to refine and personalize medical predictions further[21]. This research not only supports the use of deep learning in medical decision algorithms but also highlights its significance in evolving personalized medicine practices by providing more nuanced and individualized insights into patient outcomes.

In the processing of medical data, graph neural networks have proved to have unique advantages in analyzing graph structured data. It can effectively capture complex relational and structural information in medical data and provide more accurate support for personalized medical decisions. For example, graph neural networks can be used to analyze medical image data, mining potential patterns and features in the images to provide doctors with more accurate diagnoses.

To sum up, the research of personalized medical decision algorithm is developing in a more refined and personalized direction. As an emerging technology, graph neural network is expected to bring new breakthroughs and progress to the research and practice in this field. Therefore, the combination of graph neural network and personalized medical decision algorithm will be a potential research direction in the future medical field. The development of personalized medical decision algorithm is in a dynamic period, and the introduction of graph neural network brings a new research paradigm for this field. The combination of the two not only heralds a move towards a higher level of precision and personalization in medical decision-making.

### III. THEORETICAL BASIS

#### A. Graph neural networks

The spectral domain-based graph neural network employs a method that utilizes spectral information from graphs for feature extraction and representation learning. It examines the eigenvalues and eigenvectors of the graph's Laplacian matrix to understand its structure and features. This technique supports tasks like node and graph classification effectively. By leveraging this approach, it becomes possible to gain a deeper understanding of the graph's inherent characteristics, which aids in performing precise classification tasks that depend on spectral properties.

First with regard to the representation of graphs and the Laplacian matrix, we assume that there is a $G = (V, E)$ where $V$ is nodes and $E$ is edges. The Laplace matrix $L$ can be defined as $L = D - A$.

For Laplacian matrix $L$, its eigenvalues and eigenvectors can be obtained by spectral decomposition. Let $\lambda_1 \leq \lambda_2 \leq \cdots \leq \lambda_n$ be the eigenvalue of $L$, and the corresponding eigenvectors are $u_1, u_2, \ldots, u_n$. These eigenvectors form the spectral space of the graph $G$.

Graph convolution operations based on spectral domains can be expressed in the following form:

$$H^{(l+1)} = \sigma\left(D^{-\frac{1}{2}} A D^{-\frac{1}{2}} H^{(l)} W^{(l)}\right) \quad (1)$$

$H^{(l)}$ is the nodal eigenmatrix of the $l$ layer.

Interlayer propagation of Graph Convolutional Networks (GCN) Interlayer propagation of GCN can be expressed in the following form:

$$H^{(l+1)} = \sigma(\hat{D}^{-\frac{1}{2}} \hat{A} \hat{D}^{-\frac{1}{2}} H^{(l)} W^{(l)}) \quad (2)$$

Among them: $\hat{A} = D^{-\frac{1}{2}} A D^{-\frac{1}{2}}$ adjacency matrix is symmetric normalization. $\hat{D}$ is the degree matrix of $\hat{A}$. GCNs can be trained for graph node classification tasks using the cross-entropy loss function:

$$L = -\sum_{i=1}^{N} \sum_{k=1}^{K} y_{ik} \log(\hat{y}_{ik}) \quad (3)$$

Where: $y_{ik}$ is the label of whether the node $i$ belongs to the class $k$. $\widehat{y_{ik}}$ is the probability that the model predicts that the node $i$ belongs to the class $k$. By optimizing the loss function, the weight parameters can be used to complete the task of node classification.

While this is an overview of spectral domain based graph neural networks, recent research efforts have turned to exploring alternative matrix structures to optimize the performance of graph convolutional network (GCN) models. Among them, the adaptive graph convolutional network (AGCN) effectively captures and models the underlying structural information that was not explicitly expressed in the graph by introducing the learning distance function and the residual graph adjacency matrix. At the same time, double graph convolutional Network (DGCN) is a new approach, and a unique double graph convolutional architecture is proposed. DGCN consists of two sets of parallel graph convolution layers with shared parameters, using a normalized adjacency matrix and a matrix based on positive point mutual information. The positive point mutual information matrix captures the co-occurrence information between nodes by means of random walk. The spectral domain-based graph neural network utilizes spectral data from graphs for feature extraction and learning representations. It analyzes the eigenvalues and eigenvectors of the graph's Laplacian matrix to decode its structure and attributes. This method is effective for tasks such as node classification and graph classification. Utilizing this approach allows for a more profound comprehension of the graph's intrinsic characteristics, facilitating accurate classification tasks reliant on spectral features.

Although spectral domain graph neural networks have a solid theoretical foundation and show good performance in practical tasks, they also expose several significant limitations. Firstly, many spectral domain graph neural network methods need to decompose Laplacian matrix to obtain eigenvalues and eigenvectors in the implementation process, which often brings high computational complexity. Although ChebNet and GCN simplify this step to some extent, the entire graph is still required to be stored in memory during the calculation process, which undoubtedly consumes a lot of memory resources. Furthermore, the convolution process in spectral domain graph neural networks typically occurs on the eigenvalue matrix of the Laplacian matrix, indicating that the convolution kernel parameters do not transfer easily across different graphs. Consequently, spectral domain neural networks are generally constrained to a single graph, limiting their ability to learn across multiple graphs and to generalize. This limitation has led to fewer subsequent studies on spectral domain neural networks compared to those based in the spatial domain.

The spatial domain-based image neural network adopts the principles of traditional convolutional neural networks (CNNs) used in image processing. It expands the notion of convolution to accommodate graph data structures and defines the graph convolution operation based on the spatial correlations among nodes within the graph. As the pixels in Figure 1 constitute a two-dimensional grid structure, which can be regarded as a special form of topology diagram (see the left side of Figure 1). Similarly, when we apply the 3×3 convolution window on the image, the spatial-space-based graph convolution also simulates a similar process on the graph data. It integrates the feature sets of the central node and its neighboring nodes through convolution. This process is graphically shown on the right side of Figure 1. The core principle of spatial graph convolution lies in the propagation of node features and topological information along the edge structure of the graph, which is similar to the feature extraction and propagation of image data by CNN. In other words, spatial graph convolution can achieve the iterative updating and fusion of graph node features by simulating convolution behavior on graph data, thus playing a vital role in the analysis.

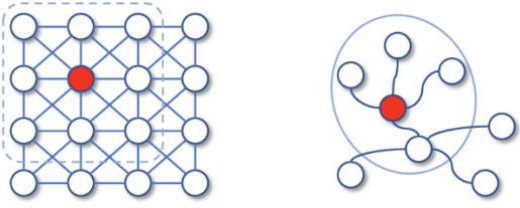

Fig. 1. Comparison of 2D convolution and graph convolution

Neural Network for Graphs (NN4G) [22] represents the inaugural implementation of spatial-domain graph neural networks in research. It employs a complex neural architecture with distinct parameters to model inter-graph relationships and to propagate information. The graph convolution technique utilized by NN4G is essentially an aggregation of neighboring node information combined with the use of a residual network to preserve the foundational data from the previous layer of the node. This operation is mathematically articulated as follows:

$$\mathbf{h}_v^{(k)} = f\left(\Theta^{(k)\bullet} \sum_{u \in N(v)} \mathbf{h}_u^{(k-1)} + \sum_{i=1}^{|N(v)|} \Theta^{(k)\bullet} \mathbf{A}_{vu} \mathbf{h}_u^{(k-1)}\right) \quad (4)$$

Where $f(\cdot)$ as the activation function, $\mathbf{h}_v^{(0)} = \mathbf{0}$. From the point of view of mathematical expression, the whole modeling process is the same as GCN. NN4G differs from GCN in that it uses an unnormalized adjacency matrix, which can result in very large differences in the scale of potential node information. The GraphSage[23] model was proposed to deal with the problem of large number of neighbors of nodes, and adopted the way of downsampling. Figure 2 illustrates how the model incorporates a sequence of aggregation functions in graph convolution to maintain consistent output regardless of node order. Specifically, GraphSage employs three symmetric aggregation functions: mean aggregation, LSTM aggregation, and pooling aggregation. The mathematical formulae utilized by the GraphSage model are as follows:

$$\mathbf{h}_v^{(k)} = \sigma\left(\mathbf{W}^{(k)} \cdot \text{agg}_k\left(\mathbf{h}_v^{(k-1)}, \{\mathbf{h}_u^{(k-1)}, \forall u \in S_N(v)\}\right)\right) \quad (5)$$

Where $\mathbf{h}_v^{(k)} = x_v$, $\text{agg}_k(\cdot)$ is the aggregation function, $S_N(v)$ is $v$ neighbor nodes of a random sample. The proposal of GraphSage has brought positive significance to the development of graph neural networks. Inductive learning makes it easier to generalize graph neural networks, while neighbor sampling leads the trend of large-scale graph learning.

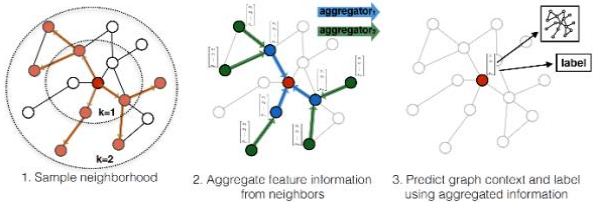

Fig. 2. GraphSage sampling and aggregation diagram

### B. Classification algorithm

The multi-scale fusion mechanism has undergone multiple stages of evolution in the development of personalized medical decision algorithms, and new fusion strategies and technologies are constantly introduced. In the early stage, multi-scale feature extraction (2010s) is based on backbone network. Researchers mainly use pre-trained classification networks [24] (such as VGG, ResNet, etc.) as backbone network to extract features of medical images. These backbone networks extract features with different scales at different levels through convolution operations.

The middle stage is the proposal of the side fusion framework (2015s). In this stage, Chen et al proposed the side fusion framework (DSS), which fuses features from deeper and shallow layers to enable the network to better obtain multi-scale feature information. New fusion strategies such as pyramid parsing module [25] (PPM) and cross feature module [26] (CFM) have been introduced into the multi-scale fusion mechanism. PPM achieves multi-scale feature output through pyramid pooling, while CFM provides cascaded feedback through selective aggregation of multi-layer features, thus further improving the feature characterization capability of the network.

The latest stage is the exploration of multi-level fusion mode (2019s till now). In the latest research, researchers began to explore the multi-level fusion mode, designed the fusion module to be recycled at each level, and constantly updated the features to make it gradually refined[27-28]. Therefore, the multi-scale fusion mechanism has undergone continuous evolution and improvement in the development of personalized medical decision algorithms. From the earliest feature extraction based on backbone network to the current multi-level fusion approach, new fusion strategies and technologies have been introduced constantly, providing more accurate and efficient support for medical image analysis and salient target detection.

At present, multi-scale fusion mechanism plays an important role in solving personalized medical decision-making problems[29]. This mechanism aims to significance target detection by integrating feature information of different scales, so as to provide more accurate support for personalized medical decision making.

However, the features extracted from each convolutional layers of these backbone networks have different scale information. The multi-scale fusion mechanism adopts a series of fusion strategies, as follows:

1. Side Fusion Framework: This method fuses information from deeper layers with shallow features through short connections, and then performs supervised learning on the merged features. This fusion strategy enables the network to obtain multi-scale feature information better. Mathematically, fusion operations can be expressed as:

$$F_{\text{fusion}} = \alpha \cdot F_{\text{deep}} + (1-\alpha) \cdot F_{\text{shallow}} \quad (6)$$

Where , ($F_{\text{fusion}}$) is the fused feature, ($F_{\text{deep}}$) and ($F_{\text{shallow}}$) are the features from the deeper and shallow layers, respectively, and ($\alpha$) is the fused weight.

2. Multilevel Fusion: In this way, the fusion module is designed to be recycled at each level, and the features are constantly updated iteratively to make them gradually refined.

3. Pyramid Parsing Module (PPM): PPM acquires feature outputs at various scales by implementing pyramid pooling at multiple levels and subsequently concatenates these channels to enhance the network's capability to gather global information. This method of pyramidal feature fusion is particularly effective in capturing multi-scale information from images, thereby enhancing the efficacy of medical image analysis.

4. Cross Feature Module (CFM) : CFM selectively aggregates multi-layer features to form a cascade feedback decoder.

The cohesive implementation of such fusion strategies empowers medical image analytics and prominent target identification models to optimally harness multi-resolution feature details, thereby furnishing enhanced precision in supporting tailored medical decision processes.

*C. Attention mechanism*

Attention mechanism [30] is a machine learning technique used by models to automatically learn and focus on important parts of input data. The attention mechanism finds broad utility across numerous domains, encompassing natural language processing (NLP) and computer vision, among others. Within NLP, it assumes a pivotal role in tasks like machine translation and text summarization. Meanwhile, in the field of computer vision, attention plays a significant role in processes such as image categorization and object recognition.

Fundamentally, the attention mechanism focuses on dynamically adjusting the model's emphasis according to the importance of elements in the input data. This allows the model to prioritize information that is more pertinent to the specific task. As depicted in Figure 3, this capability enables the model to assign different levels of importance to various parts of the input data. Consequently, this attention-based strategy makes models more flexible and accurate in processing varied inputs, thereby improving their effectiveness and precision.

Specifically, the attention mechanism is applied to different projection Spaces and different attention results are concatenated or weighted for summation. The mathematical description is as follows:

$$\text{MultiHead}(\mathbf{Q}, \mathbf{K}, \mathbf{V}) = \text{Concat}(\text{head}_1, \text{head}_2, ..., \text{head}_h)\mathbf{W}^O \quad (7)$$

which $(\text{head}_i = \text{Attention}(QW_i^Q, KW_i^K, VW_i^V))$, $(W_i^Q, W_i^K, W_i^V)$ is the projection matrix, $(W^O)$ is the output of matrix projection matrix.

The self-attention mechanism is a specialized adaptation of the attention concept, particularly designed to handle relationships within sequential data points. Here, both queries, keys, and values originate from a single input sequence. While maintaining similarity to the standard attention calculation, the distinction lies in using the identical sequence for these three components.

During the training process, the self-attention component becomes integral to the model, being optimized end-to-end. By iteratively refining the model parameters via the minimization of the loss function, the model achieves peak performance tailored to the given task. The backpropagation algorithm is usually used to update the parameters.

The attention mechanism has achieved good performance in many tasks, especially in processing sequence data and image data. This mechanism adeptly uncovers intricate connections within the data, thereby augmenting the model's capacity for representation and generalizability. Moreover, the attention mechanism boasts commendable interpretive qualities, enabling the visualization of attention weights to elucidate the model's decision-making rationale. Attention mechanism is an important machine learning technique for models to automatically learn and focus on important parts of input data.

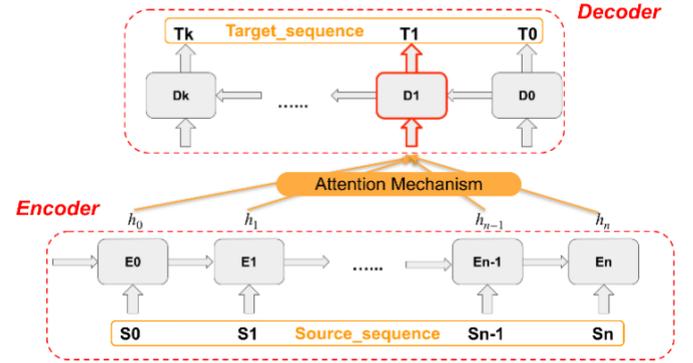

Fig. 3. Working mechanisms of the attention mechanism.

IV. PERSONALIZED MEDICAL ALGORITHM BASED ON CNN-MULTI-SCALE FUSION MECHANISM

In the study documented in this paper, the Multi-Scale Fusion CNN (MSF-CNN) tailored for Personalized Medical Decision Making incorporates a graph neural network and a multi-scale fusion approach. This MSF-CNN model utilizes convolutional neural networks at multiple scales to optimize medical decision-making on an individual basis. The integration of graph neural networks with the multi-scale fusion strategy in the MSF-CNN algorithm is intended to enhance the extraction of detailed and comprehensive features from medical imagery, facilitating the formulation of customized healthcare strategies for patients.

Firstly, through preprocessing steps, the MSF-CNN algorithm segregates medical imaging data into two distinct subsets: a training cohort and a validation ensemble, facilitating subsequent model instruction and evaluation., so as to conduct subsequent model training and evaluation. The network architecture consists of multiple convolutional layers and fusion modules, which are used to fuse features from different scales to obtain richer and more diverse feature representations.

After the feature extraction stage is completed, the extracted multi-scale features are input into the graph neural network for further processing. The purpose of graph neural networks is to capture the similarities and differences between patients by using the relationships between nodes in the graph structure, so as to make better personalized medical decisions. Subsequently, the MSF-CNN architecture undergoes training utilizing the allocated training dataset. This process employs a loss function to quantify disparities between predictions and actual outcomes, with model parameters iteratively refined through optimization methodologies, including gradient descent. The detailed execution sequence unfolds as follows:

1. Data preprocessing and partitioning: First of all, medical image data is standardized and enhanced to ensure data quality and diversity. Then, the dataset is partitioned into $(\mathcal{D}_{train})$ and $(\mathcal{D}_{test})$ according to the predetermined proportion. Where $(\mathcal{D} = \mathcal{D}_{train} \cup \mathcal{D}_{test})$ is used for learning and performance verification of the model.

2. Multi-scale fusion feature extraction: The core of the constructed MSF-CNN lies in its unique multi-scale fusion convolution layer structure. Convolution layer functions as $(f^{(l)}(\cdot))$, of which the first $(l)$ said $(l)$ layer, fusion module

($g(\cdot)$) can integrate the characteristics of different scales figure ($F_i^{(l)}$)(($i = 1,2,...,n_s$), ($n_s$) as the scale number), comprehensive features generated ($F^{(l+1)} = g\left(\sum_{i=1}^{n_s} F_i^{(l)}\right)$). In this way, multi-level medical image information from macro to micro can be captured.

3. Application of graph neural network to personalized analysis: After feature extraction, graph neural network (GNN) model, denoted as ($H(\cdot)$), is used to model the correlation between patients. In GNN, nodes represent patients or image regions, edges represent similarities or clinical associations between them, and feature updates are made through the information dissemination mechanism ($H(A,X) = \sigma(AXW)$). Where ($A$) is an adjacency matrix, ($X$) is an eigenmatrix, ($W$) represents a trainable weight matrix, while the non-linear activation function ($\sigma$) is employed to discern intricate and subtle variations among patient data.

4. Model training and optimization: $\mathcal{L} = -\sum_{i=1}^{N} y_i \log(p_i)$, ($y_i$) are real labels, ($p_i$) is a model to predict the probability of, Model parameters ($\Theta$) are adjusted by backpropagation and gradient descent (e.g. Adam) to minimize losses ($\mathcal{L}$).

5. Performance evaluation: Evaluate the performance of MSF-CNN on an independent test set ($\mathcal{D}_{test}$), with key indicators including accuracy, sensitivity, specificity and OC-ROC, etc., by comparing the agreement of the model prediction ($\hat{y}$) with the actual label ($y$). Comprehensively measure the effectiveness of algorithms in personalized medical decision making.

Ultimately, the efficacy of the trained model is assessed via examination on a distinct test dataset. Its performance in tailored medical decision-making is gauged by contrasting the discrepancies between the model's predicted outcomes and the actual classifications. Through mathematically rigorous framework design, MSF-CNN algorithm effectively combines the expressive power of deep learning with the correlation analysis of graph models to provide a more accurate and comprehensive solution for personalized medical decision-making, and can fully leverage the benefits of deep learning and graph neural network to extract more accurate and rich features and provide more accurate support for personalized medical decision-making.

## V. EXPERIMENTAL ANALYSIS

### A. Data set

We chose ISIC (International Skin Imaging Collaboration), a well-known foreign skin disease detection data set, as our experimental data set. The ISIC dataset is a large dataset dedicated to skin image analysis, containing thousands of skin microscopy images from around the world, covering many different types of skin cases. This dataset has the following advantages: The ISIC dataset covers many different types of skin cases, including but not limited to melanoma, squamous cell carcinoma, basal cell carcinoma, etc. This allows our model to be tested and evaluated in a wider range of situations; The images in the ISIC dataset have been annotated and verified by professional doctors, ensuring the accuracy and credibility of the data. Each image is accompanied by a detailed case description and pathological diagnosis information, which provides an important reference for the training and evaluation of the model.

The ISIC dataset contains thousands of high-resolution dermoscopic images covering multiple disease types and different clinical situations. This allows us to take full advantage of this data to train and test our model, thereby improving the generalization and performance of the model.

Before conducting experiments with ISIC datasets, we performed a series of data pre-processing steps to ensure the quality and consistency of the data, providing a good basis for the training and testing of the model. The dataset utilizes the Linked Data[31] methodology to consolidate various data formats, enhancing academic research by facilitating data cross-referencing and boosting interoperability among different datasets. This approach is particularly beneficial in fields like machine learning and artificial intelligence, where high-quality data is crucial for training accurate models. Linked Data also integrates diverse data sources, helping to break down data silos and enabling a more comprehensive analysis approach. This enhanced data connectivity not only improves model accuracy but also aids researchers in gaining deeper insights across multiple disciplines, such as healthcare and social sciences. First of all, the image size is unified. Since the image sizes in the ISIC data set may not be consistent, in order to ensure that the images input to the model have the same size, we adjust all images to a uniform size, usually by cropping or scaling the image to the same size. Second, we adjusted the brightness and contrast of the image. The conditions under which dermoscopic images are taken vary, and there may be differences in brightness and contrast. We use histogram equalization and other techniques to adjust the image, so that the brightness and contrast of the image are more consistent. In addition, the image is denoised. There may be various noises in dermoscopic images, such as background noise introduced by equipment or motion blur during image acquisition. In order to reduce the interference of noise to the model, we use filtering technology to denoise the image. Finally, the image is normalized. Normalization maps the pixel values of an image to a fixed range, such as $[0, 1]$ or $[-1, 1]$, to better match the input requirements of the model. This helps to improve model stability and training speed, and helps to avoid problems such as gradient explosion or gradient disappearance. Through the above data preprocessing steps, we can obtain data with higher quality and better consistency, and provide a more reliable basis for the training and testing of the model.

### B. Evaluation indicators

In the experiment of this paper, we used ISIC[32] dataset to evaluate image-text matching tasks, and adopted several common evaluation indicators to measure Recall, Precision and mAP.

Within the context of medical decision-making, the model's capability to discern positive instances is assessed using two metrics: Recall, which evaluates comprehensiveness in identifying every actual positive case, and Precision, focusing on the proportion of true positives among all instances predicted as positive:

$$\text{Recall} = \frac{\text{True Positive}}{\text{True Positive} + \text{False Negative}} \quad (8)$$

$$\text{Precision} = \frac{\text{True Positive}}{\text{True Positive} + \text{False Positive}} \quad (9)$$

Correctly identified positive instances are quantified as True Positives, whereas False Negatives denote positive scenarios inaccurately classified. Conversely, False Positives represent instances erroneously labeled as positive despite being negative.

Mean Average Precision (mAP): The mAP is utilized as an indicator to evaluate the effectiveness of a model in classification or detection tasks spanning various categories. It integrates aspects of Precision and Recall while addressing the equilibrium among distinct classes. In the context of this study, we compute the Average Precision for each category within the model and subsequently average these values to determine the overall mAP:

$$mAP = \frac{1}{N}\sum_{i=1}^{N} AP_i \qquad (10)$$

Where N is the total number of classes and $AP_i$ is the Average Precision of the I-th class. $AP_i$ is the area value under the accuracy-recall curve for that category. Through the comprehensive consideration of these evaluation indicators, we can comprehensively evaluate the algorithm's decision-making ability on the ISIC data set, and then verify its effectiveness and feasibility.

*C. Experimental setup*

A range of experiments were carried out and validated using the ISIC dataset. The experimental setup is outlined as follows: The ISIC dataset is divided into training and testing segments at a ratio of 80:20, where 80% is utilized for training the model and 20% is set aside for validation purposes. Further, we applied a 5-fold cross-validation method solely within the training segment to ensure thorough assessment of the model's performance. The architecture of the MSF-CNN model includes four convolutional layers and two pooling layers. For integration, the model employs a weighted average fusion technique, setting the fusion weights to [0.6,0.4].Regarding training specifics, the initial learning rate is set at 0.001, and is progressively decreased by a factor of 10 (to 0.0001) throughout the training phase to facilitate stable convergence. The model undergoes training over 100 epochs using a stochastic gradient descent (SGD) optimizer with a batch size of 32. Evaluation of the model is conducted using established metrics such as Precision, Recall, and Map.

Through the above experimental Settings, we can comprehensively evaluate the performance and effect of MSF-CNN algorithm in dermatological detection tasks, and provide more accurate and reliable support for personalized medical decisions.

*D. Experimental result*

In Figure 1, we designed the MSF-CNN algorithm for skin disease detection tasks, and compared its performance. The following is a detailed analysis of the comparative experimental results: Our MSF-CNN model has achieved excellent performance in dermatological detection tasks. Specifically, our model achieves 95.21%, 96.74% and 97.29% in Precision, Recall and mAP (mean precision), respectively. This shows that our MSF-CNN model achieves excellent performance in both accuracy and recall rates, and also performs well in overall average accuracy. In contrast, the ResNet model based on deep learning achieved 91.94%, 94.12% and 94.13% in Precision, Recall and mAP indexes, respectively. Although ResNet model also achieved good performance, but compared with our MSF-CNN model, there is a certain performance gap. The Precision, Recall and mAP indexes of the traditional machine learning algorithm SVM reach 90.36%, 92.89% and 93.96% respectively. Although the SVM model is slightly inferior to the deep learning model in performance, it still shows relatively good performance.

In summary, the comparison of experimental outcomes indicates that the MSF-CNN model offers considerable benefits in detecting skin diseases. This underscores the effectiveness and reliability of our personalized medicine approach, which employs a multi-scale fusion convolutional neural network. Such results provide robust support and direction for advancing personalized medical decision-making strategies in the future.

TABLE I. EXPERIMENTAL RESULTS AT DIFFERENT BASELINES

| Model | Precision | Recall | mAP |
|---|---|---|---|
| MSF-CNN | 95.21 | 96.74 | 97.29 |
| ResNet | 91.94 | 94.12 | 94.13 |
| SVM | 90.36 | 92.89 | 93.96 |

*E. Ablation experiment*

In this study, we performed a series of ablation experiments, which included comparing the MSF-CNN model with the normal CNN model. The following is a detailed analysis of the ablation experiment: Our MSF-CNN model has achieved excellent performance in the dermatological detection task. Specifically, the Precision, Recall and mAP indexes of the model reach 95.21%, 96.74% and 97.29%, respectively. This shows that the multi-scale fusion convolutional neural network we designed has significant advantages in dermatological detection tasks, recall rate and overall average accuracy. Compared with MSF-CNN model, the performance of ordinary CNN model is slightly less. Specifically, the Precision, Recall and mAP indexes of the model reach 91.32%, 90.86% and 88.21%, respectively. Compared to the MSF-CNN model, the conventional CNN model shows reduced performance in metrics such as accuracy, recall, and a general decline in mean accuracy. In conclusion, the outcomes of ablation studies reinforce the efficacy and advantage of our multi-scale fusion convolutional neural network (MSF-CNN) in tasks related to dermatological detection. Compared to the normal CNN model, our MSF-CNN model has achieved significant improvements in accuracy, recall and overall average accuracy, providing more reliable and accurate support for personalized medical decisions.

TABLE II. ABLATION RESULTS

| Model | Precision | Recall | mAP |
|---|---|---|---|
| MSF-CNN | 95.21 | 96.74 | 97.29 |
| CNN | 91.32 | 90.86 | 88.21 |

VI. CONCLUSION

This paper introduces the MSF-CNN (Multi-Scale Fusion Convolutional Neural Network for Personalized Medical Decision Making) algorithm, which achieves remarkable advancements in personalized healthcare decision support by ingeniously integrating multi-scale feature fusion within convolutional networks and the prowess of graph neural networks. The synergy of these techniques bolsters performance significantl. The primary contributions of this study are outlined as follows: multi-scale fusion strategy is used to effectively enhance the ability to capture medical image features, not only to identify macro structural features, but also to dig deep into micro details, which significantly improves the recognition accuracy of the model for complex pathological changes; The integrated graph neural network component effectively utilizes the intrinsic relationship between patients or cases, and enhances the model's ability to understand the similarities and differences of cases through graph structure analysis, which is essential for the development of highly personalized treatment plans. The experimental results showed that MSF-CNN exceeded the existing methods in many evaluation indicators, especially in the accuracy, sensitivity and specificity, which verified the effectiveness and reliability of the model in assisting doctors to develop personalized medical programs. This study not only promotes the fusion application of deep learning and graph theory methods in medical image analysis in theory, but also shows strong application potential in practice, laying a solid foundation for the future development of personalized medicine. The final conclusion is that MSF-CNN algorithm, with its unique design ideas and excellent performance, has proved its great value in the field of personalized medical decision-making. It not only improves the accuracy of diagnosis and treatment recommendations, but also provides advanced technical support to promote more personalized, efficient and safe medical care.